\documentclass[final]{siamltex}

\usepackage{graphicx, amsfonts, amsmath, wrapfig}
\usepackage{pst-sigsys,pst-plot,pst-3dplot}
\usepackage{hyperref}

\title{Random Input Sampling for Complex Models Using Markov Chain Monte Carlo}

	 \author{A. Gokcen Mahmutoglu \thanks{College of Engineering, Koc University, Sariyer-Istanbul, Turkey ({\tt amahmutoglu@ku.edu.tr}).}
	 \and Alper T. Erdogan \thanks{College of Engineering, Koc University, Sariyer-Istanbul, Turkey ({\tt alperdogan@ku.edu.tr}).}
	 \and Alper Demir \thanks{College of Engineering, Koc University, Sariyer-Istanbul, Turkey ({\tt aldemir@ku.edu.tr}).}}
\begin{document}
\maketitle

\begin{abstract}
  Many random processes can be simulated as the output of a deterministic model accepting random inputs. Such a model usually describes a complex mathematical or physical stochastic system and the randomness is introduced in the input variables of the model. When the statistics of the output event are known, these input variables have to be chosen in a specific way for the output to have the prescribed statistics. Because the probability distribution of the input random variables is not directly known but dictated implicitly by the statistics of the output random variables, this problem is usually intractable for classical sampling methods. Based on Markov Chain Monte Carlo we propose a novel method to sample random inputs to such models by introducing a modification to the standard Metropolis-Hastings algorithm. As an example we consider a system described by a stochastic differential equation (sde) and demonstrate how sample paths of a random process satisfying this sde can be generated with our technique.
\end{abstract}

\begin{keywords}
  Models with Random Input, Markov Chain Monte Carlo, Sampling Methods, Stochastic Differential Equations
\end{keywords}

\begin{AMS}
 65C20, 65C40, 65C05, 62P30
\end{AMS}

\section{Introduction}
Most algorithms employed to sample from complicated probability distributions such as rejection sampling and importance sampling assume full knowledge of the target density \cite{robert2004monte}. Contrary to these approaches Markov Chain Monte Carlo (MCMC) methods can be used to sample from  distributions for which the form of the density function is known, but the function value itself can only be evaluated up to a scalar constant.

MCMC algorithms devise a Markov chain on the sample space  of a general vector random variable. 
In typical settings the probability density function (pdf) of the distribution can only be evaluated up to a normalizing constant. 
The common Metropolis algorithm \cite{metropolis1953equation} starts with an initial state and generates samples of the random variable iteratively. 
At every step of the procedure a new state is proposed according to some proposal distribution. 
This proposal state is then accepted with a probability determined by the ratio of the pdf values for the new state and the old state. 
Because the accept-reject rule only requires the evaluation of the ratio of the probability densities for the proposed and the old state, it is sufficient to know the target pdf up to a scalar constant. 
The sole restriction of the  Metropolis algorithm is that the proposal density is symmetric and simple enough to sample directly.

One generalization of the Metropolis algorithm is the Metropolis-Hastings algorithm \cite{hastings1970monte} which can employ non-symmetric proposal densities. 
To achieve this, the acceptance probability is modified to incorporate a ratio of the proposal density values.

MCMC methods are very general tools in regard to dealing with intractable probabilistic settings. This generality allows MCMC to be integrated into many practical problems in diverse fields like computational biology \cite{huelsenbeck2001mrbayes}, statistical physics \cite{heermann1986computer}, random number generation \cite{ripley1987stochastic, devroye1986non}, artificial intelligence \cite{andrieu2003introduction} and many more. A review for the applications of MCMC can be found in \cite{neal1993probabilistic}.

\begin{figure}
  \begin{center}
		\begin{pspicture}(0,-1.5)(12,1.5)
			\psset{Alpha=135,Beta=30}
			\pnode(2,0){X} \pnode(10,0){Y} \pnode(6,0){M}
			\rput(X){\includegraphics[scale=0.3]{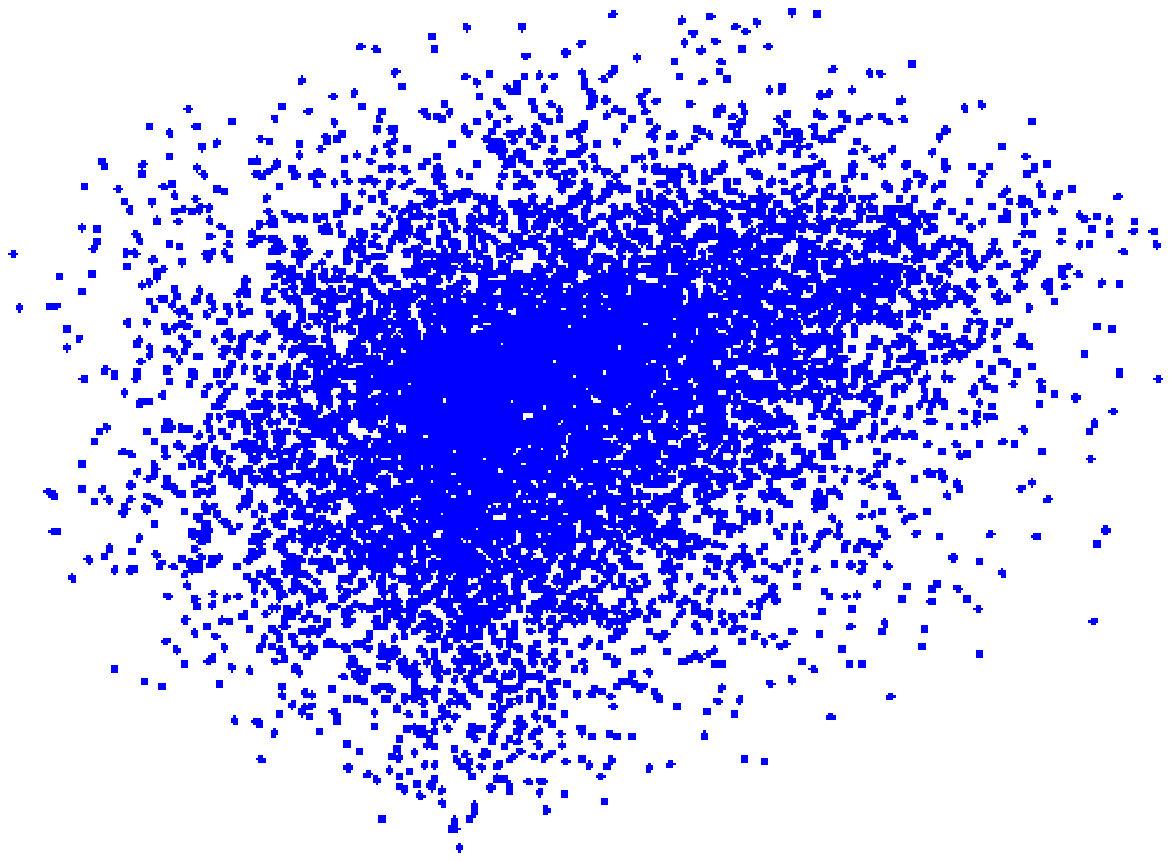}}
			\rput(X){\pstThreeDCoor[linecolor=black,linewidth=1.5\pslinewidth,nameX=$X_1$,nameY=$X_2$,nameZ=$X_3$,xMax=2.5,yMax=2.5,zMax=2,spotX=70,spotZ=0,xMin=-2,yMin=-2,zMin=-1.5,arrows=<->]}
			\rput(Y){\includegraphics[scale=0.3]{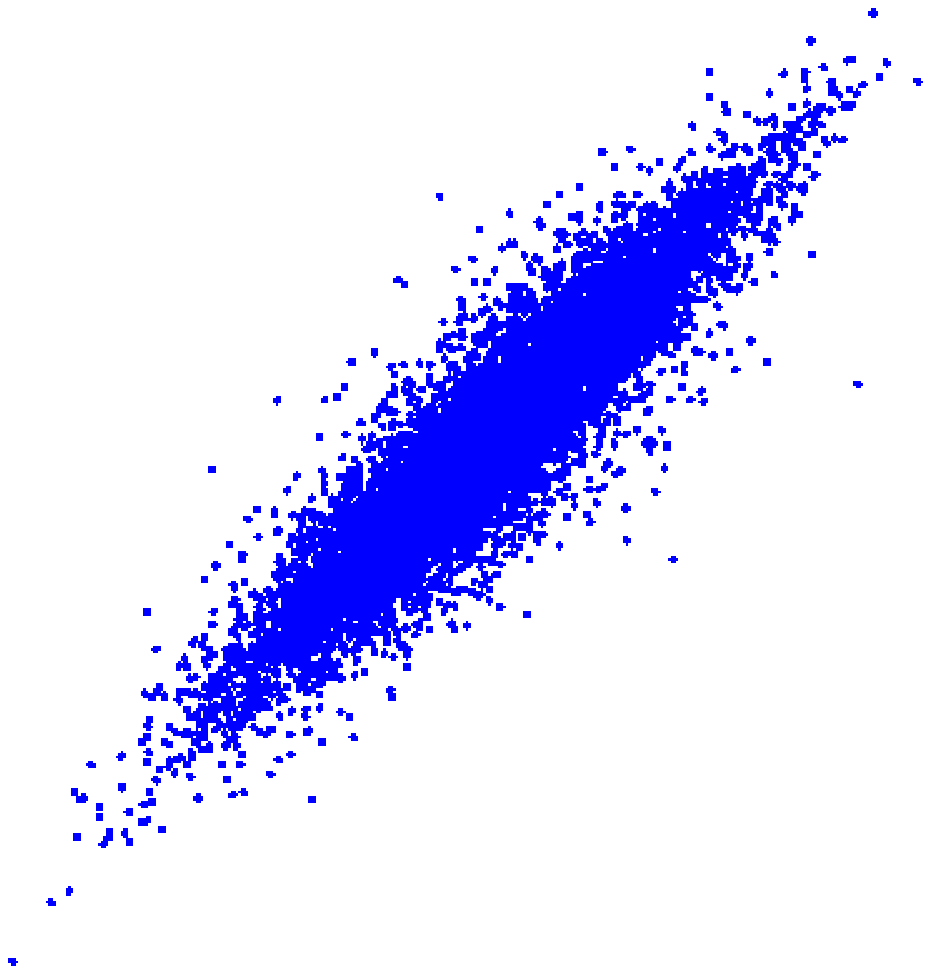}}
			\rput(Y){\psaxes[ticks=none,labels=none,arrows=->,linewidth=0.80\pslinewidth,arrows=<->]({0,0})(-1.5,-1.5)(1.5,1.5)[$Y_1$,0][$Y_2$,90]}
			\psfblock[gratioWh=1.7](M){H}{$h$}
			\pssignal([Xnodesep=1.9]X){XX}{} \ncline[arrows=->]{XX}{H}
			\pssignal([Xnodesep=-1.9]Y){YY}{} \ncline[arrows=->]{H}{YY}
		\end{pspicture}
  \end{center}
  \caption{Graphical representation of a deterministic mapping with random input and output. The output of the complex system $h$, $\mathbf{Y} = h(\mathbf{X})$, is a random vector of dimension two while the input $\mathbf{X}$ is a random vector with three dimensions.}
  \label{fig:sys}
\end{figure}

One of these broad applications deals with the model selection problem where one tries to choose a model among many competing models that is more likely to have generated the given probabilistic output data \cite{dellaportas2002bayesian}. In this paper we investigate a related problem in which the model that generates the given data is fixed, but accepts a random input with an unknown pdf. This situation arises naturally in the context of complex models which accept random inputs either as the data to be processed or as random model parameters. The realization of these models in practice can be in various ways such as lengthy and complicated computer routines, a set of involved mathematical equations or any kind of black box evaluation.
Nevertheless they can be viewed as a mapping of random variables as illustrated in figure \ref{fig:sys}.
 If $h:\mathbb{R}^n \to \mathbb{R}^m$ is a multi-valued function of multiple variables with $\mathbf{Y} = \mathbf{h}(\mathbf{X})$ and its inverse  $h^{-1}$ does not exist or cannot be computed analytically then the question arises: How does one choose $\mathbf{X}$ for $\mathbf{Y}$ to have the desired probability density $f_{\mathbf{Y}d}$?

The answer to this question is not straightforward.
First of all, generally $h$ is very complicated and all we have is some kind of routine that evaluates it for a given input. 
Therefore the unknown density of $\mathbf{X}$ can be computed through the inverse mapping only in special cases where $h$ is known explicitly and $m = n$ with the Jacobian of $h$ is globally invertible.
The use of standard sampling algorithms including MCMC to sample $\mathbf{X}$ is for this reason not possible.
Additionally more than one input distribution can generate the desired output distribution creating an issue of non-uniqueness.

 To address this problem we first review some Markov chain and MCMC theory then we provide a detailed description of the problem at hand and a toy example to demonstrate the concept before proceeding to develop a solution. Finally we conclude our discussion with a numerical example of a stochastic differential equation and demonstrate how our method can be used to sample from the space of solution paths to this equation.

 \section{Background} \label{sec:background}
In this section we present some elementary Markov chain and MCMC theory which will be required for later discussion. 

 \begin{definition}
   A sequence of indexed random variables $ \mathbf{X}_i, \; i \in \mathbb{N}$ is called a Markov chain if the following property holds for every measurable set $\mathcal{A} \subseteq \mathbb{R}^n$
   \begin{equation}\mathrm{Pr}(\mathbf{X}_{i+1} \in \mathcal{A} | \mathbf{X}_{i}, \mathbf{X}_{i-1}, \dots, \mathbf{X}_0) = \mathrm{Pr}(\mathbf{X}_{i+1} \in \mathcal{A} | \mathbf{X}_{i})\;.\end{equation}
   $T(\mathbf{x}, \mathcal{A}) = \mathrm{Pr}(\mathbf{X}_{i+1} \in \mathcal{A} | \mathbf{X}_i = \mathbf{x})$ is called its transition kernel with the transition density $\tau(\mathbf{x},\mathbf{x}')$, where
   \begin{equation}T(\mathbf{x}, \mathcal{A}) = \int \limits_{\mathcal{A}} \tau(\mathbf{x}, \mathbf{x}')\mathrm{d}\mathbf{x}' , \qquad \mathbf{x} \in \mathbb{R}^n\end{equation}
 \end{definition}

 \begin{definition}\label{def:stat_dist}
   $f(\mathbf{x})$ is called a stationary distribution of the Markov chain if it satisfies
   \begin{equation}f(\mathbf{x}') = \int \limits_{\mathbb{R}^n} f(\mathbf{x})\tau(\mathbf{x},\mathbf{x}') \mathrm{d}\mathbf{x} \;.\end{equation}

 \end{definition}

 \begin{lemma}\label{lem:det_bal}
   A sufficient condition for $f(\mathbf{x})$ to be a stationary distribution of the Markov chain $\mathbf{X}_i$ is the detailed balance equation:
   \begin{equation}
     f(\mathbf{x})\tau(\mathbf{x},\mathbf{x}') = f(\mathbf{x}')\tau(\mathbf{x}',\mathbf{x}), \qquad \mathbf{x},\mathbf{x}' \in \mathbb{R}^n\;.
     \label{eqn:det_bal}
   \end{equation}
   
   \begin{proof}
     Integrating both sides of equation (\ref{eqn:det_bal}) and using Definition \ref{def:stat_dist},
     \begin{align}
       \int \limits_{\mathbb{R}^n} f(\mathbf{x})\tau(\mathbf{x},\mathbf{x}') \mathrm{d}\mathbf{x} &= \int \limits_{\mathbb{R}^n} f(\mathbf{x}')\tau(\mathbf{x}',\mathbf{x}) \mathrm{d}\mathbf{x} \nonumber \\
       &= f(\mathbf{x}')\int \limits_{\mathbb{R}^n} \tau(\mathbf{x}',\mathbf{x}) \mathrm{d}\mathbf{x} \nonumber \\
       &= f(\mathbf{x}')
       \label{eqn:det_bal_proof}
     \end{align} \qquad
   \end{proof}
 \end{lemma}

 Note that Lemma \ref{lem:det_bal} gives a sufficient condition, and the necessary condition is much looser \cite{mira2000non}. Furthermore under certain conditions the stationary distribution is unique \cite{athreya1996convergence}.

 Above definitions and Lemma \ref{lem:det_bal} are sufficient to describe the Metropolis-Hastings (and the Metropolis algorithm as its special case) in a formal way. 
 Given a target distribution $f(\mathbf{x})$ for the random vector $\mathbf{X}$ the strategy of the algorithm is to construct a Markov chain on the state space of interest, $\mathbb{R}^n$, and choose a transition kernel such that the Markov chain has  $f(\mathbf{x})$ as its stationary distribution. 
 This is accomplished in two stages. At the first stage the procedure takes a random step in the state space according to some proposal density $p(\mathbf{x},\mathbf{x}')$ which describes the probability of moving from the state, $\mathbf{X}_i = \mathbf{x}$, to the next one, $\mathbf{X}_{i+1} = \mathbf{x}'$. Most common choice for $p$ uses a form of increment on $\mathbf{x}$ such that $\mathbf{x}' = \mathbf{x} + \Delta\mathbf{x}$. Commonly used densities for the random increment $\Delta\mathbf{x}$ are tractable ones like the uniform and the Gaussian density.
 At the second stage of the algorithm a decision is made whether the chain will advance to $\mathbf{x}'$ as its next state or stay at $\mathbf{x}$. The decision mechanism uses the ratio $\frac{f(\mathbf{x}')}{f(\mathbf{x})}$ in the decision rule

 \begin{equation}
   \alpha(\mathbf{x},\mathbf{x}') = \mathrm{min}\left(1, \frac{f(\mathbf{x}')}{f(\mathbf{x})} \frac{p(\mathbf{x}',\mathbf{x})}{p(\mathbf{x},\mathbf{x}')}\right)
   \label{eqn:accpt-rejct}
 \end{equation}
 
 \noindent which gives the acceptance probability of the proposed move. After evaluating the accept-reject ratio, a random number $u$ is sampled according to a standard uniform distribution and the move is accepted if $u \leq \alpha(\mathbf{x},\mathbf{x}')$. If the proposed state is not accepted the Markov chain remains in its previous state.
 Theorem \ref{thrm:met-has} shows that the distribution of the samples taken in this way indeed converges to $f(\mathbf{x})$.

 \begin{theorem} \label{thrm:met-has}
   The transition kernel of the Metropolis-Hastings algorithm satisfies the detailed balance condition and $f(\mathbf{x})$ is the stationary distribution of the resulting Markov chain.
   \begin{proof}
     The transition kernel $T(\mathbf{x}, \mathcal{A})$ can be written as the sum of two probabilities: The probability of an accepted step to a point $\mathbf{x}'$ in $\mathcal{A}$ and the probability of a rejection while the point $\mathbf{x}$ lies in $\mathcal{A}$.
     \begin{equation*}
       T(\mathbf{x},\mathcal{A}) = \int \limits_{\mathcal{A}} p(\mathbf{x},\mathbf{x}') \alpha(\mathbf{x},\mathbf{x}') \mathrm{d}\mathbf{x}' + 1_{\{\mathbf{x} \in \mathcal{A}\}} \int \limits_{\Omega} p(\mathbf{x},\mathbf{x}')(1- \alpha(\mathbf{x},\mathbf{x}')) \mathrm{d}\mathbf{x}'
     \end{equation*}

     \noindent Hence the transition density is given by
     \[
     \tau(\mathbf{x},\mathbf{x}') = p(\mathbf{x},\mathbf{x}')\alpha(\mathbf{x},\mathbf{x}') + \delta_{\mathbf{x}}(\mathbf{x}') r(\mathbf{x}) \;,
     \]
     \noindent where $\delta_{\mathbf{x}}(\mathbf{x}')$ is the point mass at $\mathbf{x}$ and $r(\mathbf{x}) = 1 - \int_{\Omega} \alpha(\mathbf{x},\mathbf{x}')p(\mathbf{x},\mathbf{x}') \mathrm{d}\mathbf{x}'$ is the probability that the chain does not leave its current position $\mathbf{x}$.

     Lemma \ref{lem:det_bal} gives us a way of checking whether this transition kernel has the desired pdf as its stationary distribution. If we now check if equation (\ref{eqn:det_bal}) is satisfied we find for the first summand of the transition density 
     \begin{align}
       f(\mathbf{x})p(\mathbf{x},\mathbf{x}')\alpha(\mathbf{x},\mathbf{x}') &= f(\mathbf{x})p(\mathbf{x},\mathbf{x}')\mathrm{min}\left(1,\frac{f(\mathbf{x}')}{f(\mathbf{x})} \frac{p(\mathbf{x}',\mathbf{x})}{p(\mathbf{x},\mathbf{x}')}\right) \nonumber\\
       &= \mathrm{min}\left(f(\mathbf{x})p(\mathbf{x},\mathbf{x}'),f(\mathbf{x}')p(\mathbf{x}',\mathbf{x}) \right) \nonumber\\
       &= \mathrm{min}\left(\frac{f(\mathbf{x})p(\mathbf{x},\mathbf{x}')}{f(\mathbf{x}')p(\mathbf{x}',\mathbf{x})}, 1\right) f(\mathbf{x}')p(\mathbf{x}',\mathbf{x}) \nonumber\\
       &=  f(\mathbf{x}')p(\mathbf{x}',\mathbf{x})\alpha(\mathbf{x}',\mathbf{x}) \quad.
       \label{eqn:prf_det_bal}
     \end{align}
    \noindent Finally for the second summand the requirement is trivially satisfied
     \[
     \delta_{\mathbf{x}}(\mathbf{x}') r(\mathbf{x}) = \delta_{\mathbf{x}'}(\mathbf{x}) r(\mathbf{x}') \quad,
     \]
     \noindent and this completes the proof. \qquad
   \end{proof}
   
 \end{theorem}

 Further discussion of Markov chain and MCMC theory is outside the scope of this paper but excellent material on this subject can be found in \cite{tierney1994markov} and \cite{tierney1998note}.
 
\section{An Illustrative Example}

Now let us recap the problem described in the introduction.
Suppose we are given a general many-to-one, non-isometric map $h: \mathbb{R}^n \to \mathbb{R}^m$ which maps a random vector $\mathbf{X}$ to another random vector $\mathbf{Y} = h(\mathbf{X})$ where $\mathbf{X} \in \mathbb{R}^n$, $\mathbf{Y} \in \mathbb{R}^m$ and let $f_{\mathbf{Y}d}$ be the desired probability density of $\mathbf{Y}$. 
Given $f_{\mathbf{Y}d}$ how must $f_{\mathbf{X}}$ be chosen such that the transformed variable $\mathbf{Y} = h(\mathbf{X})$ has the desired pdf?

Given this setting one might be tempted to construct a Markov chain in the space of the input variables to sample $\mathbf{X}$ while evaluating the accept-reject rule probabilities of the Metropolis-Hastings algorithm in the space of the random output vector $\mathbf{Y}$.

As the following toy-example illustrates, this method does not result in a Markov chain in the space of input variables with the desired stationary distribution $f_{\mathbf{Y}d}$ of the output variables.

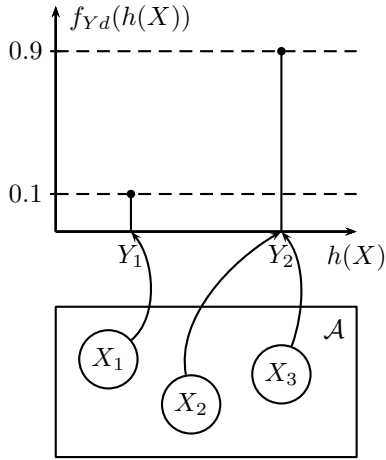
\begin{wrapfigure}{r}{55mm}
  \begin{center}
		\begin{pspicture}(-0.5,-3)(4,3)
			\psaxes[ticks=none,labels=none,arrows=->](0,0)(0,0)(4,3)[$h(X)$,-90][$f_{Yd}(h(X))$,-25]
			\psframe[dimen=middle](0,-3)(4,-1)
			\psTick{0}(0,0.5) \uput[180](0,0.5){$0.1$}
			\psTick{0}(0,2.4) \uput[180](0,2.4){$0.9$}
			\psline[linestyle=dashed](0,0.5)(4,0.5)
			\psline[linestyle=dashed](0,2.4)(4,2.4)
			\pnode(1,0){Y1} \pnode(3,0){Y2}
			\uput[-90](Y1){$Y_1$} \uput[-90](Y2){$Y_2$}
			\psline(Y1)([offset=0.5]Y1) \pscircle*([offset=0.5]Y1){2\pslinewidth}
			\psline(Y2)([offset=2.4]Y2) \pscircle*([offset=2.4]Y2){2\pslinewidth}
			\cnodeput{0}(0.7,-1.7){X1}{$X_1$} \nccurve[angleA=40,angleB=-50]{->}{X1}{Y1}
			\cnodeput{0}(1.8,-2.3){X2}{$X_2$} \nccurve[angleA=100,angleB=-150]{->}{X2}{Y2}
			\cnodeput{0}(3,-1.9){X3}{$X_3$} \nccurve[angleA=70,angleB=-50]{->}{X3}{Y2}
			\uput[-90](3.7,-1){$\mathcal{A}$}
		\end{pspicture}
  \end{center}
  \caption{A toy-example to illustrate the problem of mapping the state variables $X$ to another random variable $Y$ with the desired probability distribution $f_{Yd}$.}
  \label{fig:toy-example}
\end{wrapfigure}

Figure \ref{fig:toy-example} describes a discrete state space consisting of three states $\mathcal{A} = \{X_1, X_2, X_3\}$.  
The arrows represent a many-to-one function $h: \mathcal{A} \to \mathcal{B}$ with $h(X_1) = Y_1$ and $h(X_2) = h(X_3) = Y_2$ and $Y_1,Y_2 \in \mathcal{B}$. 
It is of no importance if $\mathcal{B}$ is discrete or continuous but the range of $h$ is discrete for obvious reasons. 

The desired distribution of $Y$ is $f_{Yd}(Y_1) = 0.9$ and $f_{Yd}(Y_2) = 0.1$.  
 The results about Markov chains and the Metropolis-Hastings algorithm given in section \ref{sec:background} can easily be adopted to general finite state spaces and to this specific example.

Now suppose that we are running the Metropolis algorithm on $\mathcal{A}$ with a symmetric proposal distribution $P$. For the current state $X_i$ a new state is proposed according to the rule
\begin{equation*}
  \mathrm{Pr}(X_j|X_i) = P(X_i,X_j) = 
  \begin{cases}
    \frac{1}{2} & \text{if} \; i \neq j \\
    0 & \text{if} \; i = j
  \end{cases} \quad.
\end{equation*}

\noindent Together with the accept-reject rule of the Metropolis algorithm, this results in a Markov chain with the transition probabilities $T(X_i,X_j) = P(X_i,X_j) \mathrm{min}(1,\frac{f_{Yd}(h(X_j))}{f_{Yd}(h(X_i))})$ for $i \neq j$ and the transition probability matrix 

\begin{equation}
  \mathbf{T} = \left(\begin{array}{ccc}
    \frac{16}{18} & \frac{1}{18} & \frac{1}{18} \\
    \frac{1}{2} & 0 & \frac{1}{2} \\
    \frac{1}{2} & \frac{1}{2} & 0  		
  		
  		\end{array}\right)
  \label{eq:transMat}
\end{equation}
The left eigenvector of this matrix that corresponds to the eigenvalue $1$ gives us the stationary distribution of the chain, which is $(\frac{9}{11}, \frac{1}{11}, \frac{1}{11}) $. 
It can be easily seen that this distribution does not provide the desired stationary distribution on the range of $h$. 
In fact this distribution corresponds to a function $h'$  which maps $X_3$ to a different value $h'(X_3)$ which has the same probability as $h'(X_2)$. This behavior can also be observed with the more general Metropolis-Hastings algorithm by choosing a non-symmetric proposal distribution and applying the corresponding accept-reject rule. 

This toy-example illustrates clearly that to address the problem of creating the target probability density $f_{\mathbf{Y}d}$ we have to take the properties of the mapping $h$ into account and modify the Metropolis-Hastings algorithm accordingly.

\section{Modification of MCMC with a Probing Term}

The reason that the above example fails to converge to the desired stationary distribution $f_{Yd}$ lies within the properties of the general mapping $h$. First $h$ is not one-to-one and hence the probability of a state $Y_i$ appearing in the chain on $\mathcal{B}$ depends on the probability of all the states $X_j$ on the space of inputs for which $h(X_j) = Y_i$ holds. Additionally for the continuous case, even if $h$ was one-to-one it would not necessarily be an isometry  so that volumes are distorted under the mapping creating a similar effect on the stationary distribution. In this section we develop a method to overcome the shortcomings of MCMC sampling for the problem described in the previous section.

In this context for the general case we first implement a probing procedure for the mapping $h$ by using the output distribution that results when the input parameters are sampled uniformly and independently. Then we show that a modification of the target density with this uniform output density can be used in the space of parameters for the accept-reject rule in MCMC to achieve the desired density $f_{\mathbf{Y}d}$ on the range of $h$. 

\begin{theorem}\label{thrm:compensation}
  Let $\mathbf{U}$ be a uniform random vector on the probability space $(\Omega,\; \mathcal{F},\; F_{\mathbf{U}})$ where $\Omega$ is a bounded subset of $\mathbb{R}^n$ such that $f_{\mathbf{U}}(\mathbf{u}_1) = f_{\mathbf{U}}(\mathbf{u}_2)$ for all $\mathbf{u}_1, \mathbf{u}_2 \in \Omega$ with $f_{\mathbf{U}}$ as the probability density function of the cumulative distribution function $F_{\mathbf{U}}$. 
  And let $h:\mathbb{R}^n \to \mathbb{R}^m$ be a mapping satisfying the required regularity conditions such that $(\Omega', \; \mathcal{F}')$ with $\Omega' = h(\Omega)$ is the induced sample space by $h$ and the associated $\sigma\text{-algebra}$. 
  Then a random variable $\mathbf{X} \in \mathbb{R}^n$ constructs another random variable $\mathbf{Y} = h(\mathbf{X}), \; \mathbf{Y} \in \mathbb{R}^m$ with the desired probability density $f_{\mathbf{Y}d}$ if $\mathbf{X}$ has the unnormalized probability density 

 \[
 f_{\mathbf{X}}(\mathbf{x}) \propto \frac{f_{\mathbf{Y}d}(h(\mathbf{x}))}{f_{\mathbf{Q}}(h(\mathbf{x}))}
 \]
  
 \noindent where $f_{\mathbf{Q}}$ is the probability density of the transformed random variable $\mathbf{Q} = h(\mathbf{U})$.

  \begin{proof}

    Consider the bounded sample space $\Omega \subseteq \mathbb{R}^n$, $\Omega = [\alpha_1, \; \beta_1] \times [\alpha_2, \; \beta_2] \times \dots \times [\alpha_n, \; \beta_n]$ in which we assume $f_{\mathbf{Q}}$ is strictly positive. 
    For the cumulative distribution function of $\mathbf{Q}$ we get 
  
\begin{equation}
  F_{\mathbf{Q}}(\mathbf{q}) = \mathrm{Pr}(\mathbf{Q} \leq \mathbf{q}) = \mathrm{Pr}(\{\mathbf{u}: \mathbf{u} \in \Omega, \; h(\mathbf{u}) \leq \mathbf{q}\}) \quad.
  \label{eqn:prob_dist_q}
\end{equation}

\noindent which can be written with the indicator function as

\begin{equation}
  F_{\mathbf{Q}}(\mathbf{q}) = \int \limits_{\Omega} 1_{\{\mathbf{u}: h(\mathbf{u}) \leq \mathbf{q}\}} f_{\mathbf{U}}(\mathbf{u}) \mathrm{d}\mathbf{u} 
  \label{eqn:prob_dist_q_ind}
\end{equation}

\noindent Note that the indicator function in equation (\ref{eqn:prob_dist_q_ind}) can be expressed with the components of the random vector $\mathbf{q}$ and the function $h$ as

\begin{equation}
  1_{\{\mathbf{u}: h(\mathbf{u}) \leq \mathbf{q}\}} = s(q_1 - h_1(\mathbf{u})) s(q_2 - h_2(\mathbf{u}))  \dots s(q_m - h_m(\mathbf{u}))
  \label{eqn:indicator_function}
\end{equation}

\noindent where $s$ is the unit step function.

The pdf of $\mathbf{Q}$ is given by $f_{\mathbf{Q}}(\mathbf{q}) = \frac{\partial^m F_{\mathbf{Q}}(\mathbf{q})}{\partial q_1 \partial q_2 \dots \partial q_m}$. Using generalized functions and equation (\ref{eqn:indicator_function}) we can write this expression as

\begin{eqnarray}
  f_{\mathbf{Q}}(\mathbf{q}) &=& \int \limits_{\Omega} \delta(q_1 - h_1(\mathbf{u})) \delta(q_2 - h_2(\mathbf{u}))  \dots \delta(q_m - h_m(\mathbf{u})) f_{\mathbf{U}}(\mathbf{u}) \mathrm{d}\mathbf{u} \nonumber\\
  &\propto& \int \limits_{\alpha_1}^{\beta_1} \int \limits_{\alpha_2}^{\beta_2} \dots \int \limits_{\alpha_n}^{\beta_n} \delta(q_1 - h_1(\mathbf{u})) \delta(q_2 - h_2(\mathbf{u})) \dots \delta(q_m - h_m(\mathbf{u}))  \mathrm{d}\mathbf{u}
  \label{eqn:pdf_q}
\end{eqnarray}

If we now set the distribution of the input random variable $\mathbf{X}$ proportional to the ratio of the desired distribution of $\mathbf{Y}$ and the distribution of $\mathbf{Q}$,

\begin{equation}
  f_{\mathbf{X}}(\mathbf{x}) \propto \frac{f_{\mathbf{Y}d}(h(\mathbf{x}))}{f_{\mathbf{Q}}(h(\mathbf{x}))}
  \label{eqn:compensated_dist_x}
\end{equation}

\noindent we have for the cumulative distribution function of $\mathbf{Y} = h(\mathbf{X})$ 

\begin{eqnarray}
  F_{\mathbf{Y}}(\mathbf{y}) &=& \mathrm{Pr}(\mathbf{Y} \leq \mathbf{y}) \nonumber\\
  &=& \mathrm{Pr}(\{\mathbf{x}: \mathbf{x} \in \Omega, \; h(\mathbf{\mathbf{x}}) \leq \mathbf{y}\})\nonumber \\
  &=& \int \limits_{\Omega} 1_{\{\mathbf{x}: h(\mathbf{x}) \leq \mathbf{y}\}} f_{\mathbf{X}}(\mathbf{x}) \mathrm{d}\mathbf{x} \nonumber \\
  &=& \int \limits_{\Omega} 1_{\{\mathbf{x}: h(\mathbf{x}) \leq \mathbf{y}\}} \frac{f_{\mathbf{Y}d}(h(\mathbf{x}))}{f_{\mathbf{Q}}(h(\mathbf{x}))} \mathrm{d}\mathbf{x} \quad. 
  \label{compensated_dist_y}
\end{eqnarray}

\noindent Finally we have for the output probability density,

\begin{eqnarray}
  f_{\mathbf{Y}}(\mathbf{y}) &=& \frac{\partial^m}{\partial y_1 \partial y_2 \dots \partial y_m}F_{\mathbf{Y}}(\mathbf{y}) \nonumber \\
  &\propto& \int \limits_{\alpha_1}^{\beta_1} \int \limits_{\alpha_2}^{\beta_2} \dots \int \limits_{\alpha_n}^{\beta_n} \delta(y_1 - h_1(\mathbf{x})) \delta(y_2 - h_2(\mathbf{x})) \dots \delta(y_m - h_m(\mathbf{x}))  \frac{f_{\mathbf{Y}d}(h(\mathbf{x}))}{f_{\mathbf{Q}}(h(\mathbf{x}))} \mathrm{d}\mathbf{x} \nonumber \\
  &\propto& \frac{f_{\mathbf{Y}d}(\mathbf{y})}{f_{\mathbf{Q}}(\mathbf{y})} \underbrace{\int \limits_{\Omega} \delta(\mathbf{y} - h(\mathbf{x})) \mathrm{d}\mathbf{x}}_{f_\mathbf{Q}(\mathbf{y})} \nonumber \\
  &\propto& f_{\mathbf{Y}d}(\mathbf{y})
  \label{eqn:compensated_density_y} 
\end{eqnarray}

\noindent Since both are normalized probability densities $f_\mathbf{Y} = f_{\mathbf{Y}d}$ holds and the distribution of the image of samples on $\mathbb{R}^n$ will be equal to the desired distribution on $\mathbb{R}^m$. \qquad
\end{proof}
\end{theorem}

Theorem \ref{thrm:compensation} shows that we can find the distribution of a random variable $\mathbf{X}$ that gives us the desired density $f_{\mathbf{Y}d}$ through the mapping $h$ provided that we know the uniform input distribution $f_\mathbf{Q}$. This can be accomplished by modifying the Metropolis-Hastings accept-reject rule in (\ref{eqn:accpt-rejct}) as

\begin{equation}
  \alpha(\mathbf{x},\mathbf{x}') = \mathrm{min}\left(1, \frac{f_{\mathbf{Y}d}(h(\mathbf{x}'))}{f_{\mathbf{Y}d}(h(\mathbf{x}))} \frac{p(\mathbf{x}',\mathbf{x})}{p(\mathbf{x},\mathbf{x}')}
  \frac{f_{\mathbf{Q}}(h(\mathbf{x}))}{f_{\mathbf{Q}}(h(\mathbf{x}'))} \right) \quad.
  \label{eq:mod-accpt-rejct}
\end{equation}

Note that Theorem \ref{thrm:compensation} assumes a bounded support for the uniformly sampled random vector $\mathbf{Q}$, with $\Omega = [\alpha_1, \; \beta_1] \times [\alpha_2, \; \beta_2] \times \dots \times [\alpha_n, \; \beta_n]$. 
This assumption implies that the support of the input vector $\mathbf{X}$ is equal to or a subset of $\Omega$. Therefore in case $\mathbf{X}$ has unbounded support, this technique will sample a truncated version of the input random vector. Nevertheless practical difficulties caused by this fact can be overcome with an adjustment of $\Omega$ which theoretically can be chosen arbitrarily large. 

Furthermore Theorem \ref{thrm:compensation} gives us only the unnormalized pdf which is sufficient to sample $\mathbf{X}$ with MCMC. But the above method can be used irrespective of the specific sampling method once this density is normalized. Hence we obtain a general method to control the input of complex systems with prescribed random outputs. 

In practical applications one will not always be able to compute $f_{\mathbf{Q}}$ analytically. In these situations $f_{\mathbf{Q}}$ will have to be substituted with an approximation $\hat{f}_{\mathbf{Q}}$. For this purpose one can use various density estimation schemes available. For large data sets nonparametric schemes like kernel density estimators and nearest neighbour methods \cite{bishop2006pattern} can be used. For other settings Bayesian schemes like the EM algorithm \cite{redner1984mixture} can be employed for inference.

\section{An Application: Stochastic Differential Equations}

In this section we demonstrate an example for our algorithm on stochastic differential equations. In this case the model is given by a differential equation driven by random noise and the input random variable takes the form of the solution to this equation.

Consider the one dimensional It\={o} stochastic differential equation

\begin{align}
  &\mathrm{d}X_t = b(X_t,t)\mathrm{d}t + a(X_t,t)\mathrm{d}W_t \;, \qquad 0 \leq t \leq T 
  \label{eqn:sde_gen} \\
  &X_0 = c \nonumber
\end{align}

\noindent where $a, b: \mathbb{R} \times [0, T] \to \mathbb{R}$ are measurable functions and $W_t$ is the Wiener process. 

A numerical treatment of this equation can be done by discretization using the simple Euler scheme.

\begin{equation}
  X_{i+1} = X_i + b(X_i, t_i)\Delta t + a(X_i, t_i) \Delta W_i \;, \qquad 0 = t_0 \leq t_1 \leq \dots \leq t_N = T 
  \label{eqn:dsc_sde}
\end{equation}

We now set $\Delta X_i = X_{i} - X_{i-1}$ and define the random vectors \linebreak $\Delta \mathbf{X} = (\Delta X_1, \Delta X_2 , \dots , \Delta X_N)^T$ and $\mathbf{Y} = h(\Delta \mathbf{X}) = (Y_1, Y_2, \dots, Y_N)^T$ with

\begin{equation}
  Y_i = \frac{\Delta X_i - b(X_{i-1}, t_{i-1})\Delta t}{a(X_{i-1}, t_{i-1})} = \Delta W_{i-1} 
  \label{eqn:sde_Y}
\end{equation}

Note that $\Delta \mathbf{X}$ together with $X_0$ completely determines the sample path. Hence if we can sample $\Delta \mathbf{X}$ such that it satisfies equation (\ref{eqn:dsc_sde}), that means we can generate a solution path to the stochastic differential equation. The distribution of $\Delta \mathbf{X}$ is unknown but we know that $\mathbf{Y}$ is a Gaussian random vector with i.i.d.~zero mean components with variance $\Delta t$. Using this fact we can employ the modified MCMC algorithm to sample solution paths. 

For this general class of stochastic differential equations we can obtain the pdf of the output vector when the input random variables are sampled uniformly. First we derive the expression of the joint output distribution for the uniformly sampled input variables. Let $\Delta \mathbf{U} \in \mathbb{R}^N$ be a random vector with i.i.d.~components distributed uniformly in $[-\rho,\rho]$ and $\mathbf{Q} = h(\Delta \mathbf{U}) \in \mathbb{R}^N$ another random vector with the joint pdf $f_{\mathbf{Q}}$. We can express $f_{\mathbf{Q}}$ as the product of conditional pdfs as follows.

\begin{equation}
  f_{\mathbf{Q}}(\mathbf{q}) = f_{\mathbf{Q}}(q_1,q_2,\dots,q_N) = f(q_1)f(q_2|q_1)\dots f(q_N|q_{N-1},q_{N-2},\dots,q_1)
  \label{eqn:sde_uni_pdf}
\end{equation}

It can easily be seen from equation (\ref{eqn:sde_Y}) that each of these conditional pdfs are uniform in a range determined by the previous values of $q_i$. Particularly since 

\[
Q_i = \frac{\Delta U_i - b(U_{i-1},t_{i-1})\Delta t}{a(U_{i-1},t_{i-1})} \;, \qquad U_k = U_{k-1} + \Delta U_k
\]
\noindent we have

\begin{align}
  f(q_i|q_{i-1},q_{i-2},\dots,q_1) &= f(q_i|u_{i-1},u_{i-2},\dots,u_1,u_0) \nonumber\\
  &\propto |a_{i-1}|\left[s\Big(q_i - \frac{-\rho - b_{i-1}}{|a_{i-1}|}\Big) - s\Big(q_i - \frac{\rho - b_{i-1}}{|a_{i-1}|}\Big)\right]
\end{align}

\noindent where $a_k = a(u_{k}, t_{k})$, $b_k = b(u_k,t_k)$ and $s$ is the step function.
Now we can write the joint density function.

\begin{equation}
  f_{\mathbf{Q}}(q_1,q_2,\dots,q_N) \propto 
  \begin{cases}
    \prod \limits_{i = 0}^{N-1} |a_i| & \text{if} \; q_i \in \left[\frac{-\rho - b_{i-1}}{|a_{i-1}|}, \frac{\rho - b_{i-1}}{|a_{i-1}|}\right] \forall i \in \{1,2,\dots,N\}\\
    0 & \text{o.w.}
  \end{cases}
  \label{eqn:sde_uni_joint}
\end{equation}

As discussed in the previous section, the restriction of $\Delta U_i$ in $[-\rho, \rho]$ is a practical necessity and does not create any problems in real world applications since $\rho$ can be chosen arbitrarily large, and the points where $f_{\mathbf{Q}}$ is zero can be viewed as proposals of impossible states and rejected immediately. 

Combining  equations (\ref{eq:mod-accpt-rejct}) and (\ref{eqn:sde_uni_joint}) the whole accept-reject probability of the MCMC algorithm can be written as

\begin{equation}
  \alpha(\Delta \mathbf{x}, \Delta \mathbf{x}') = \mathrm{min}\left(1, \frac{f_{\mathbf{Y}d}(h(\Delta \mathbf{x}'))}{f_{\mathbf{Y}d}(h(\Delta \mathbf{x}))} \frac{f_{\mathbf{Q}}(h(\Delta \mathbf{x}))}{f_{\mathbf{Q}}(h(\Delta \mathbf{x}'))} \right)
  \label{eqn:sde_accpt_rej} 
\end{equation}

\noindent with $f_{\mathbf{Y}d}(\mathbf{y}) \propto e^{-\mathbf{y}^T \mathbf{y} / (2\Delta t)}$.

As a numerical example for the above procedure consider the linear stochastic differential equation

\begin{align} 
  \label{eqn:sde_gbm}
  & \mathrm{d}X_t = \mu X_t \mathrm{d}t + \sigma X_t \mathrm{d}W_t \;, \qquad 0 \leq t \leq 1 \\ 
  & X_0 = 1 \nonumber
\end{align}

\noindent where $\mu$ and $\sigma$ are scalar constants. This equation describes the geometric Brownian motion \cite{oksendal2003stochastic} which finds applications in mathematical finance, particularly in the Black-Scholes model of financial markets \cite{black1973pricing}.
This is a good example for demonstration purposes because one can obtain its solution analytically.
A stochastic process satisfying equation (\ref{eqn:sde_gbm}) will have the form

\begin{equation}
  X_t = X_0 e^{(\hat{\mu}t + \sigma W_t)}
\end{equation}

\noindent where $\hat{\mu} = \mu - \frac{\sigma^2}{2}$. It's pdf has a lognormal distribution,

\begin{equation}
  f_{X_{t}}(x,t) = \frac{1}{\sigma x \sqrt{2 \pi t}} e^{-(\ln{x} - \ln{X_0}) - \hat{\mu}t)^2/(2\sigma^2 t)} 
  \label{eqn:gbm_pdf}
\end{equation}

\noindent and the autocorrelation of $X_t$ is given by

\begin{equation}
  R(s,t) = e^{\mu(s+t)}(e^{\sigma^2 \min{(s,t)}}-1) \quad.
  \label{eqn:gbm_acf}
\end{equation}

\begin{figure}
  \begin{center}
    \includegraphics[width=9cm]{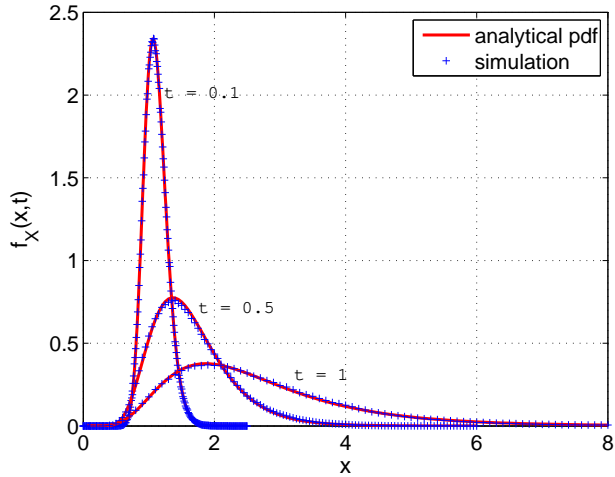}
  \end{center}
  \caption{Analytical probability density functions of $X_t$ at t = 0.1, t= 0.5 and t=1 compared with the empirical pdfs of the simulation data.}
  \label{fig:gbm_pdf}
\end{figure}

\begin{figure}
  \begin{center}
    \includegraphics[width=9cm]{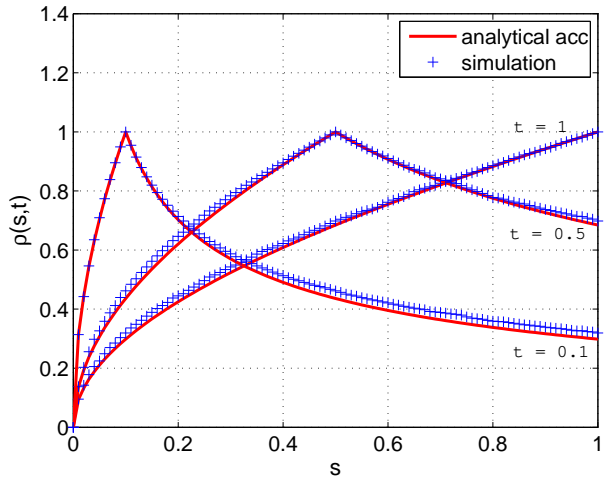}
  \end{center}
  \caption{Normalized autocorrelation of $X_t$ at three different time points compared with simulation data.}
  \label{fig:gbm_corr}
\end{figure}

Figures \ref{fig:gbm_pdf} and \ref{fig:gbm_corr} display the results of a simulation with the modified MCMC algorithm. The scalar constants in equation (\ref{eqn:sde_gbm}) were chosen as $\mu = 1$ and $\sigma = 0.5$. The time axis was divided in one hundred equal length intervals with $\Delta t = 0.01$. Initially $5.1 \times 10^6$ samples were generated and the first $10^5$ samples were discarded as the burn-in length. For this setting $\rho$ was chosen to be 2 and a uniform distribution in $[-0.2,0.2]$ was used as the proposal distribution for $\Delta \mathbf{X}$. Figure \ref{fig:gbm_pdf} shows three analytical pdfs at different time points compared with the empirical pdfs obtained from the simulation data and figure \ref{fig:gbm_corr} shows the normalized autocorrelation with one time point held fixed and the second one varied between 0 and 1. These graphical results verify that the sample paths built using our algorithm converge to the desired stationary distribution and hence satisfy the given stochastic differential equation.

One noteworthy property of numerical solutions using the modified MCMC algorithm is that all points of a sample path get sampled in parallel as opposed to classical iterative methods such as the Euler-Maruyama scheme \cite{kloeden1994stochastic}. These methods usually begin with the initial value $X_0$, and sample later points of the solution path with an iterative update rule given by the difference equation (\ref{eqn:dsc_sde}). For this reason dealing with more complicated settings like stochastic boundary value problems of the form  

\begin{align}
  & \mathrm{d}X_t = b(X_t,t)\mathrm{d}t + a(X_t,t)\mathrm{d}W_t \;, \qquad 0 \leq t \leq T 
  \label{eqn:sde_bound} \\
  & h(X_0,X_T) = 0 \nonumber
\end{align}

\noindent becomes troublesome because the points of a sample path are not independent of its future values.
On the other hand the incorporation of boundary conditions to the modified MCMC algorithm is straightforward since the points of the proposed sample paths are obtained simultaneously with independent increments. 

\section{Conclusion}

We have presented a solution to the problem of input variable sampling for complex stochastic models with prescribed output distribution. This approach is based on a modification to the Metropolis-Hastings algorithm with an additional expression which can be viewed as a probing term for the model of interest. Our algorithm is easy to implement, benefits from the extensive literature on MCMC and hence we believe that it can be adapted to a variety of applications. We have demonstrated one such application on general stochastic differential equations viewing them from the perspective of stochastic input-output models enabling us to apply our algorithm to obtain solution paths. 

Although this paper is based on MCMC, the approach taken to tackle the input variable sampling problem does not require any specific sampling method to be used. The algorithm presented here can be implemented equally well with other sampling methods once the output distribution for uniformly sampled input variables is worked out and therefore offers a fresh approach for dealing with general stochastic models.


\begin{thebibliography}{10}

\bibitem{andrieu2003introduction}
{\sc C.~Andrieu, N.~De~Freitas, A.~Doucet, and M.I. Jordan}, {\em An
  introduction to mcmc for machine learning}, Machine learning, 50 (2003),
  pp.~5--43.

\bibitem{athreya1996convergence}
{\sc K.B. Athreya, H.~Doss, and J.~Sethuraman}, {\em On the convergence of the
  markov chain simulation method}, The Annals of Statistics, 24 (1996),
  pp.~69--100.

\bibitem{bishop2006pattern}
{\sc C.M. Bishop and SpringerLink~(Service en~ligne)}, {\em Pattern recognition
  and machine learning}, vol.~4, Springer New York, 2006.

\bibitem{black1973pricing}
{\sc F.~Black and M.~Scholes}, {\em The pricing of options and corporate
  liabilities}, The journal of political economy,  (1973), pp.~637--654.

\bibitem{dellaportas2002bayesian}
{\sc P.~Dellaportas, J.J. Forster, and I.~Ntzoufras}, {\em On bayesian model
  and variable selection using mcmc}, Statistics and Computing, 12 (2002),
  pp.~27--36.

\bibitem{devroye1986non}
{\sc L.~Devroye}, {\em Non-uniform random variate generation},  (1986).

\bibitem{hastings1970monte}
{\sc W.K. Hastings}, {\em Monte carlo sampling methods using markov chains and
  their applications}, Biometrika, 57 (1970), p.~97.

\bibitem{heermann1986computer}
{\sc D.W. Heermann}, {\em Computer simulation methods: in theoretical physics},
   (1986).

\bibitem{huelsenbeck2001mrbayes}
{\sc J.P. Huelsenbeck and F.~Ronquist}, {\em Mrbayes: Bayesian inference of
  phylogenetic trees}, Bioinformatics, 17 (2001), pp.~754--755.

\bibitem{kloeden1994stochastic}
{\sc P.E. Kloeden, E.~Platen, and H.~Schurz}, {\em Stochastic differential
  equations}, Numerical Solution of SDE Through Computer Experiments,  (1994),
  pp.~63--90.

\bibitem{metropolis1953equation}
{\sc N.~Metropolis, A.W. Rosenbluth, M.N. Rosenbluth, A.H. Teller, E.~Teller,
  et~al.}, {\em Equation of state calculations by fast computing machines}, The
  journal of chemical physics, 21 (1953), p.~1087.

\bibitem{mira2000non}
{\sc A.~Mira and C.J. Geyer}, {\em On non-reversible markov chains}, Monte
  Carlo methods, 26 (2000), p.~95.

\bibitem{neal1993probabilistic}
{\sc R.M. Neal and University of~Toronto. Department~of Computer~Science}, {\em
  Probabilistic inference using Markov chain Monte Carlo methods}, Citeseer,
  1993.

\bibitem{oksendal2003stochastic}
{\sc B.K. {\O}ksendal}, {\em Stochastic differential equations: an introduction
  with applications}, Springer Verlag, 2003.

\bibitem{redner1984mixture}
{\sc R.A. Redner and H.F. Walker}, {\em Mixture densities, maximum likelihood
  and the em algorithm}, SIAM review,  (1984), pp.~195--239.

\bibitem{ripley1987stochastic}
{\sc B.D. Ripley and Ebooks Corporation}, {\em Stochastic simulation}, vol.~21,
  Wiley Online Library, 1987.

\bibitem{robert2004monte}
{\sc C.P. Robert and G.~Casella}, {\em Monte Carlo statistical methods},
  Springer Verlag, 2004.

\bibitem{tierney1994markov}
{\sc L.~Tierney}, {\em Markov chains for exploring posterior distributions},
  the Annals of Statistics, 22 (1994), pp.~1701--1728.

\bibitem{tierney1998note}
\leavevmode\vrule height 2pt depth -1.6pt width 23pt, {\em A note on
  metropolis-hastings kernels for general state spaces}, Annals of Applied
  Probability,  (1998), pp.~1--9.

\end{thebibliography}
\end{document}